\documentclass[letterpaper, 10 pt, conference]{ieeeconf}  

 \usepackage{times}
\usepackage{epsfig}
\usepackage{graphicx}
\usepackage{amsmath}
\usepackage{amssymb}

\usepackage{caption}
\usepackage{subcaption}
\usepackage{wrapfig}
\usepackage{tabularx}
\usepackage{placeins}
\usepackage{dblfloatfix}
\usepackage[table]{xcolor}

\usepackage{algorithm}
\usepackage{algpseudocode}
\usepackage{pifont}

\usepackage{amsmath}
\usepackage{amssymb}
\usepackage{amsfonts}
\usepackage{mathtools}
\usepackage{bm}
\usepackage{mathrsfs}
\usepackage[table]{xcolor}
\usepackage{color}
\usepackage{verbatim}

\definecolor{LightRed}{rgb}{0.92,0.92,0.92}
\def\x{\mathbf{x}}

\def\y{\mathbf{y}}

\IEEEoverridecommandlockouts                              

\title{Learning a Domain-Invariant Embedding for Unsupervised Domain Adaptation Using Class-Conditioned Distribution Alignment 
}

\author{Alexander  J. Gabourie$^{1}$, Mohammad Rostami$^{2}$, Philip E. Pope$^{3}$, \\ Soheil Kolouri$^{3}$, and Kuyngnam Kim$^{3}$
\thanks{$^{1}$Alexander J. Gabourie
 is with the Department of Electrical Engineering,
        Stanford University, Stanford, CA, USA
        {\tt\small ajgabourie@hrl.com}}%
\thanks{$^{2}$Mohammad Rostami is with the Department of Electrical and Systems Engineering, University of Pennsylvania, Philadelphia, PA, US. {\tt\small mrostami@seas.upenn.com}}
\thanks{ Philip Pope, Soheil Kolouri, and Kyungnam Kim are with Information and Systems Laboratory, HRL Laboratories, LLC,
        Malibu, CA, USA
        {\tt\small {  pepope, skolouri, kkim }@hrl.com}}%
}

\begin{document}

\maketitle
\thispagestyle{empty}
\pagestyle{empty}

\begin{abstract}

We address the problem of \textit{unsupervised domain adaptation} (UDA) by learning a   cross-domain agnostic embedding space, where the distance between  the probability distributions of the two  source and  target visual domains is minimized. We use the output space of a shared cross-domain deep encoder to model the embedding space and
 use the Sliced-Wasserstein Distance (SWD) to measure and minimize the distance between the embedded distributions of two source and target domains to enforce the embedding to be domain-agnostic.
 Additionally, we use the source domain labeled data to train a deep classifier from the embedding space to the label space to enforce the embedding  space to be discriminative. 
 As a result of this training scheme, we  provide an effective solution to train  the deep classification network   on the source domain such that it will generalize  well on the target domain, where only   unlabeled training data is accessible. To mitigate the challenge of class matching, we also align corresponding classes  in the embedding space   by using high confidence pseudo-labels for the target domain, i.e. assigning the class for which the source classifier has a high prediction probability. We provide  
 experimental results on UDA benchmark tasks to demonstrate that our method is effective and leads to state-of-the-art performance.

\end{abstract}

\section{Introduction}
Deep learning classification algorithms have surpassed performance of humans for a wide range of computer vision applications. However, this achievement is conditioned on availability of high-quality labeled datasets to supervise training  deep neural networks. Unfortunately, preparing huge labeled datasets  is not feasible for many situations as data labeling and annotation can be expensive~\cite{rostami2018crowdsourcing}.
Domain adaptation~\cite{glorot2011a} is a paradigm   to address the problem of  labeled data scarcity in computer vision, where the goal is to improve learning speed and model generalization as well as to avoid expensive redundant model retraining. The major idea   is to overcome  labeled data scarcity in a target domain  
by transferring knowledge   from  a related auxiliary source domain,  where labeled data is  easy and cheap to obtain.

 A common technique in domain adaptation literature  is to embed data from the two source and   target visual domains in an   intermediate
 embedding space such that common cross-domain discriminative relations are captured in the embedding space.  For example, 
  if the data from   source and target domains   have similar class-conditioned probability distributions   
  in the embedding space, then a  classifier  trained solely    using   labeled data from the source domain will generalize well on  data points that are drawn from
  the target domain distribution~\cite{redko2017theoretical,rostami2019deep}.  
  
 In this paper, we propose a novel unsupervised adaptation (UDA) algorithm following the above explained   procedure.
 Our approach is a simpler, yet  effective,  alternative for adversarial learning techniques that have been more dominant   to address probability matching indirectly for UDA~\cite{tzeng2017adversarial,yi2017dualgan,motiian2017few}.
 Our contribution is two folds. First, we train the shared encoder by minimizing the Sliced-Wasserstein Distance (SWD)~\cite{rabin2011bwasserstein}  between the source and the target distributions in the embedding space. We also  train  a classifier network simultaneously
 using the  source domain labeled data. A major benefit of SWD  over alternative probability metrics is that it can be computed efficiently. Additionally, SWD is known to be suitable for   gradient-based optimization which is essential for deep  learning~\cite{redko2017theoretical}.   
 Our second contribution is to circumvent the class matching challenge~\cite{saito2017asymmetric} by  minimizing  SWD between conditional distributions in sequential iterations for better performance compared to prior UDA methods that match probabilities explicitly. At each iteration, we assign  pseudo-labels only to the target domain data that the classifier predicts the assigned class label with high probability and use this portion of target data to minimize the SWD between conditional distributions. As more learning iterations are performed, the number of target data points with correct  pseudo-labels grows and progressively enforces distributions to align class-conditionally. We provide 
 experimental results on  benchmark problems, including ablation and sensitivity studies, to demonstrate that our method is effective.
 
\section{Background and Related Work}     
There are two major approaches in the literature to address domain adaption. The approach for a   group of methods is based on  preprocessing the target domain data points. The target data is  mapped from the target domain to the source domain such that the target data structure is preserved in the source~\cite{saenko2010adapting}.
Another  common approach is to map data from both domains to 
a latent domain invariant space~\cite{daume2009frustratingly}.  Early   methods within the second approach  learn a linear subspace as the invariant space ~\cite{gopalan2011domain,gong2012geodesic}   where the target domain data points distribute similar to the source domain data points. A linear subspace is not suitable for capturing complex distributions. For this reason,  recently deep neural networks have been used to model  the intermediate space as the output of the network. The network is trained such that the source and the target domain distributions in its output possess minimal discrepancy. Training procedure can be done both by adversarial learning~\cite{goodfellow2014generative} or directly minimizing the distance between the two distributions~\cite{courty2017optimal}.

Several important UDA methods use adversarial learning.
  Ganin et al.~\cite{ganin2016domain} pioneered and developed an effective method to match two distributions indirectly by using
  adversarial learning.   Liu et al.~\cite{liu2016coupled}  
and Tzeng et al.~\cite{tzeng2017adversarial} use the Generative Adversarial Networks (GAN) structure~\cite{goodfellow2014generative} to tackle domain adaptation. The idea is to train two competing (i.e., adversarial) deep neural networks to match the source and the target distributions. A generator network  maps data points from both domains to 
the domain-invariant space and a binary discriminator network is trained to classify the data points, with each domain considered as a class, based on the representations of the target and the source data points. The generator  network is trained such that eventually the discriminator cannot distinguish between the two domains, i.e. classification rate becomes $50\%$.  

A second group of domain adaptation algorithms
  match   the distributions directly in the embedding   by using a shared cross-domain mapping such that  the distance between the two distributions is minimized with respect to a   distance metric~\cite{courty2017optimal}.   Early methods use simple metrics such the Maximum Mean Discrepancy (MMD)  for this purpose~\cite{gretton2009covariate}. MMD  measures  the distances between the distance between distributions simply as the distance between the mean   of embedding features. In contrast, more recent techniques that use a shared deep encoder, employ the Wasserstein metric~\cite{villani2008optimal} to address UDA~\cite{courty2017optimal,damodaran2018deepjdot}. Wasserstein metric is shown to be a more accurate probability metric   and can be minimized effectively by deep learning first-order optimization techniques.  A major benefit of matching distributions directly   is existence of theoretical guarantees. In particular,
Redko et al.~\cite{redko2017theoretical} provided theoretical guarantees for using a Wasserstein metric to address domain adaptation. Additionally, adversarial learning often
requires deliberate  architecture engineering, optimization initialization, and selection of hyper-parameters to be stable~\cite{roth2017stabilizing}. In some cases, adversarial learning also suffers from a phenomenon known as mode collapse~\cite{metz2016unrolled}. That is, if the data distribution is a multi-modal distribution, which is the case for most classification problems, the generator network  might not generate samples from some modes of the distribution.   These challenges are easier to address when the distributions are matched directly.

As Wasserstein distance is finding more applications in deep learning, efficient computation of
 Wasserstein distance has become an active area of research. The reason is that Wasserstein distance is defined in form of  a linear programming optimization and solving this optimization problem  is computationally expensive for high-dimensional data. Although computationally efficient variations and approximations of the Wasserstein distance have been recently proposed \cite{cuturi2013sinkhorn,solomon2015convolutional,oberman2015efficient}, these variations still require an additional optimization in each iteration of the stochastic gradient descent (SGD) steps to match distributions.  Courty et al.~\cite{courty2017optimal} used a regularized version of the optimal transport for domain adaptation. Seguy et al.~\cite{seguy2017large} used a dual stochastic gradient
algorithm for solving the regularized optimal transport problem. Alternatively, we propose to address the above challenges using Sliced Wasserstein Distance (SWD). Definition of SWD is motivated by the fact that
in contrast to higher dimensions,
the Wasserstein distance for one-dimensional distributions has a closed form solution which can be computed efficiently. This fact  is used to approximate Wasserstein distance by  SWD, which is a computationally efficient approximation   and has recently drawn interest from the machine learning and computer vision communities~\cite{rabin2011bwasserstein,bonneel2015sliced, carriere2017sliced,deshpande2018generative,csimcsekli2018sliced}.

\section{Problem Formulation} 
Consider a source domain, $\mathcal{D}_{\mathcal{S}}=(\bm{X}_{\mathcal{S}}, \bm{Y}_{\mathcal{S}})$, with $N$ labeled samples, i.e. labeled images, where $\bm{X}_{\mathcal{S}}=[\bm{x}_1^s,\ldots,\bm{x}_N^s]\in\mathcal{X}\subset\mathbb{R}^{d\times N}$ denotes the samples
and $\bm{Y}_{\mathcal{S}}=[\y^s_1,...,\y^s_N]\in \mathcal{Y}\subset\mathbb{R}^{k\times N}$ contains the corresponding labels. Note that label $\y^s_n$ identifies the membership of $\bm{x}^s_n$ to one or multiple of the $k$ classes (e.g. digits $1,...,10$ for hand-written digit recognition). We assume that the source samples are drawn i.i.d. from the source joint probability distribution, i.e. $(\bm{x}_i^s,\bm{y}_i)\sim p(\bm{x}^{\mathcal{S}},y)$. We denote the source marginal distribution over $\bm{x}^{\mathcal{S}}$ with $p_{S}$.   Additionally, we  have a related target domain (e.g. machine-typed digit recognition)  with $M$ unlabeled  data points  $\bm{X}_{\mathcal{T}}=[\bm{x}_1^t,\ldots,\bm{x}_M^t]\in\mathbb{R}^{d\times M}$. Following existing UDA methods, we assume that the same type of labels in the source domain holds for the target domain. The target samples are drawn from the target marginal distribution $\bm{x}_i^t\sim  p_{T}$. We also know that despite similarity between these domains, distribution discrepancy exists between these two domains, i.e. $p_{S}\neq p_{T}$. Our goal is to classify the unlabeled target data points through knowledge transfer from the source domain. 
Learning a good classifier for the source data points is a straight forward problem as given a large enough number of source samples, $N$, a parametric function $f_{\theta}:\mathbb{R}^d\rightarrow \mathcal{Y}$, e.g., a deep neural network with concatenated learnable parameters $\theta$, can be trained to map samples to their corresponding labels using standard supervised learning solely in the source domain. The training is conducted via minimizing the empirical risk, $\hat{ \theta}=\arg\min_{\theta}\hat{e}_{\theta}=\arg\min_{\theta}\sum_i \mathcal{L}(f_{\theta}(\bm{x}_i^s),\bm{y}_i^s)$, with respect to a proper loss function,  $\mathcal{L}(\cdot)$ (e.g., cross entropy loss). The learned classifier $f_{\hat{\theta}}$ generalizes well on testing data points if they are drawn from the training data point's distributions. Only then, the empirical risk is a suitable surrogate for the real risk function, $e = \mathbb{E}_{(\bm{x},y)\sim p(\bm{x}^{\mathcal{S}},y^{\mathcal{S}})}(\mathcal{L}(f_{\theta}(\bm{x}),y))$.   Hence the naive approach of using $f_{\hat{\theta}}$ on the target domain might not be effective as given the discrepancy between the source and target distributions,  $f_{\hat{\theta}}$  might  not generalize well on the target domain. Therefore, there is a need for adapting the training procedure of $f_{\hat{\theta}}$ by incorporating unlabeled target data points such that the learned knowledge from the source domain could be transferred and used for classification   in the target domain using only the unlabeled samples.

The main challenge is to circumvent the problem of discrepancy between the source and the target domain distributions. To that end,  
the mapping $f_\theta(\cdot)$ can be decomposed into a feature extractor $\phi_{\bm{v}}(\cdot)$  and a classifier $h_{\bm{w}}(\cdot)$, such that $f_\theta = h_{\bm{w}}\circ \phi_{\bm{v}}$, where $\bm{w}$ and $\bm{v}$ are the corresponding learnable parameters, i.e. $\theta=(\bm{w},\bm{v})$. 
The core idea is  to learn the   feature extractor function, $\phi_{\bm{v}}$, for both domains such that the domain specific distribution of the extracted features to be similar to one another. 
The feature extracting function $\phi_{\bm{v}}: \mathcal{X}\rightarrow \mathcal{Z}$, maps the data points from both domains to an intermediate embedding space $\mathcal{Z}\subset \mathbb{R}^f$ (i.e., feature space) and the classifier  $h_{\bm{w}}: \mathcal{Z}\rightarrow \mathcal{Y}$ maps the data points representations in the embedding space to the label set. Note that, as a deterministic function, the feature extractor function $\phi_{\bm{v}}$ can change the distribution of the data in the embedding. Therefore, if $\phi_{\bm{v}}$ is learned such that the discrepancy between the source and target distributions is minimized in the embedding space, i.e., discrepancy between $p_{\mathcal{S}}(\phi(\bm{x}^s))$ and $p_{\mathcal{T}}(\phi(\bm{x}^t))$ (i.e., the embedding is domain agnostic), then  the classifier would generalize well on the target domain and could be used to label the target domain data points. This is the core idea behind various prior domain adaptation approaches in the literature \cite{murez2018image,ghifary2016deep}. 

\section{Proposed Method}

\begin{figure}[t!]
    \centering
    \includegraphics[width=\linewidth]{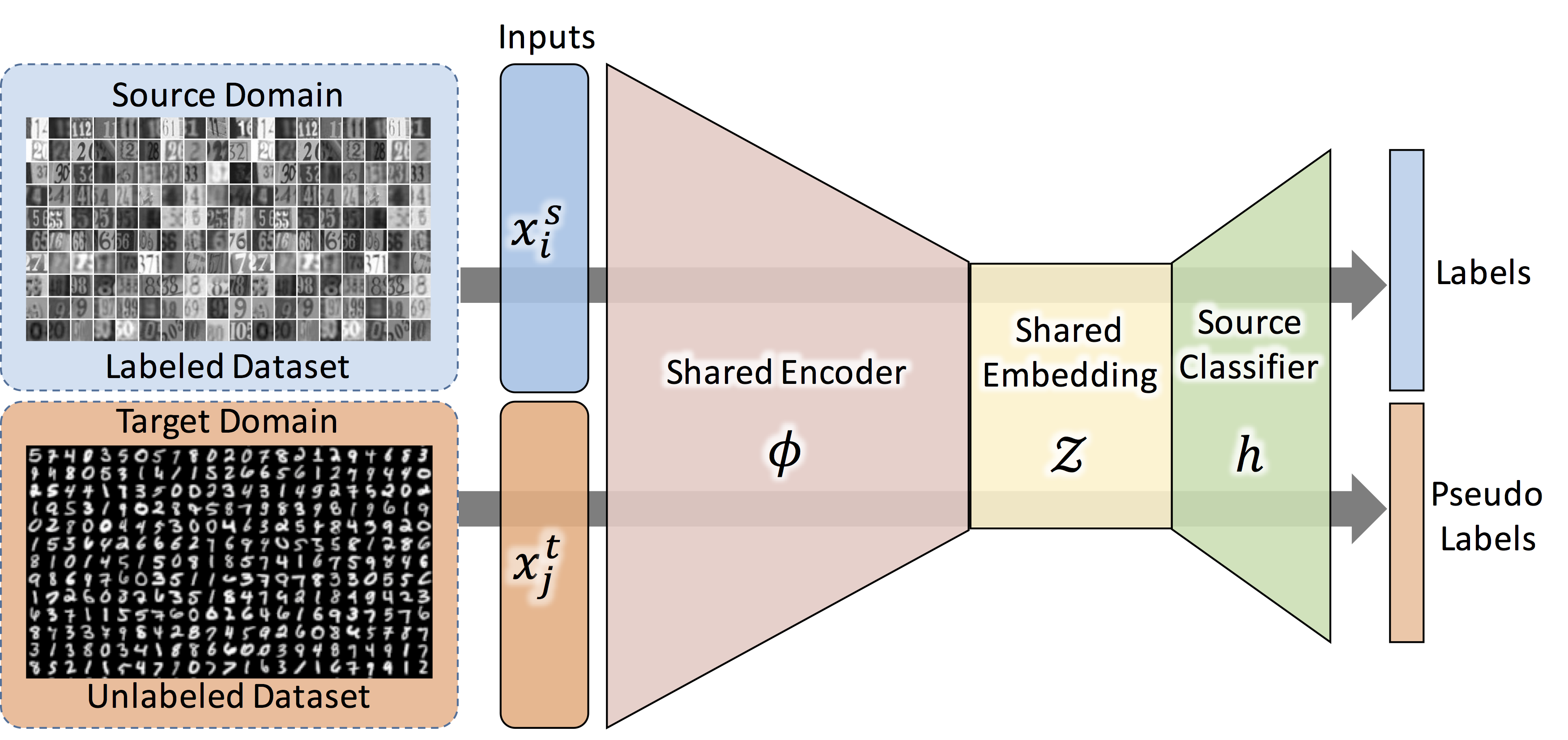}
         \caption{Architecture of the proposed unsupervised domain adaptation framework.}
         \label{figDL:LL}
\end{figure}

We consider the case where the feature extractor, $\phi_{\bm{v}}(\cdot)$, is a deep convolutional encoder with weights $\bm{v}$ and the classifier $h_{\bm{w}}(\cdot)$ is a  shallow fully connected neural network with weights $\bm{w}$. The last layer of the classifier network is a softmax layer that assigns  a membership probability distribution   to any given data point. It is often the case that   the labels of data points are assigned according to the class with maximum predicted probability.  In short, the encoder network is learned to mix  both domains such that the extracted features in the embedding are: 1)  domain agnostic in terms of data distributions, and 2) discriminative for the source domain to make learning $h_{\bm{w}}$ feasible.  Figure~\ref{figDL:LL} demonstrates system level presentation of our framework. Following this framework, the UDA reduces to solving the following optimization problem to solve for $\bm{v}$ and $\bm{w}$:
\begin{equation}
\begin{split}
\min_{\bm{v},\bm{w}}& \sum_{i=1}^N \mathcal{L}\big(h_{\bm{w}}(\phi_{\bm{v}}(\bm{x}_i^s)),\bm{y}_i^s\big)\\&+\lambda D\big(p_{\mathcal{S}}(\phi_{\bm{v}}(\bm{X}_{\mathcal{S}})),p_{\mathcal{T}}(\phi_{\bm{v}}(\bm{X}_{\mathcal{T}}))\big),
\end{split}
\label{eq:mainPrMatch}
\end{equation}    
where $D(\cdot,\cdot)$ is a discrepancy measure between the probabilities and $\lambda$ is a trade-off parameter. The first term in Eq.~\eqref{eq:mainPrMatch} is empirical risk for classifying the source labeled data points from the embedding space and the second term is the cross-domain probability matching loss. The encoder's learnable parameters are learned using data points from both domains and the   classifier parameters are simultaneously  learned using  the source domain labeled data. 

A major remaining question is to select a proper   metric. First, note that the actual distributions  $p_\mathcal{S}(\phi(\bm{X}_\mathcal{S}))$ and $p_\mathcal{T}(\phi(\bm{X}_\mathcal{T}))$  are   unknown and we can rely only on observed samples from these distributions. Therefore, a sensible discrepancy measure, $D(\cdot,\cdot)$, should be able to measure the dissimilarity between these distributions only based on the drawn samples.  In this work, we use the SWD~\cite{rabin2011wasserstein} as it is computationally efficient to compute SWD from drawn samples from the corresponding distributions. 
 More importantly, the SWD is a good approximation for the optimal transport~\cite{bonnotte2013unidimensional} which   has gained interest in deep learning community as it  is an effective distribution metric and its gradient is non-vanishing.  


The idea behind the SWD is to project two $d$-dimensional probability distributions into their marginal one-dimensional distributions, i.e., slicing the high-dimensional distributions, and to approximate the Wasserstein distance by integrating the Wasserstein distances between the resulting marginal probability distributions over all possible one-dimensional subspaces.
For the distribution $p_\mathcal{S}$,  a one-dimensional slice of the distribution is defined as:
\begin{equation}
\mathcal{R}p_\mathcal{S}(t;\gamma)=\int_\mathcal{S} p_\mathcal{S}(\bm{x})\delta(t-\langle\gamma, \bm{x}\rangle)d\bm{x},
\label{eq:radon}
\end{equation}
where $\delta(\cdot)$ denotes the Kronecker delta function,  $\langle \cdot ,\cdot\rangle$ denotes the vector dot product, $\mathbb{S}^{d-1}$ is the $d$-dimensional unit sphere and $\gamma$ is the projection direction. In other words, $\mathcal{R}p_\mathcal{S}(\cdot;\gamma)$ is a marginal distribution of $p_\mathcal{S}$  obtained from integrating $p_\mathcal{S}$ over the hyperplanes orthogonal to $\gamma$. The SWD then can be computed by integrating the Wasserstein distance between sliced distributions  over all $\gamma$:
\begin{eqnarray}
SW(p_\mathcal{S},p_\mathcal{T})=   \int_{\mathbb{S}^{d-1}} W(\mathcal{R} p_\mathcal{S}(\cdot;\gamma),\mathcal{R} p_\mathcal{T}(\cdot;\gamma))d\gamma
\label{eq:radonSWDdistance}
\end{eqnarray}
 where $W(\cdot)$ denotes the Wasserstein distance.
The main advantage of using the SWD is that, unlike the Wasserstein distance, calculation of the SWD does not require a numerically expensive optimization. This is due to the fact that the Wasserstein distance between two one-dimensional probability distributions has a closed form solution and  is equal to the $\ell_p$-distance between the inverse of their cumulative distribution functions 
Since only samples from distributions are available, the  one-dimensional Wasserstein distance can be approximated as the $\ell_p$-distance between the sorted samples~\cite{rostami2019sar}. 
The integral in Eq.~\eqref{eq:radonSWDdistance} is approximated using a Monte Carlo style numerical integration.  Doing so, the SWD between $f$-dimensional samples $\{\phi(\x_i^\mathcal{S})\in \mathbb{R}^f\sim p_\mathcal{S}\}_{i=1}^M$ and $\{\phi(\x_i^\mathcal{T})\in \mathbb{R}^f \sim p_\mathcal{T}\}_{j=1}^M$ can be approximated as the following sum:
{\small
\begin{equation}
SW^2(p_\mathcal{S},p_\mathcal{T})\approx \frac{1}{L}\sum_{l=1}^L \sum_{i=1}^M| \langle\gamma_l, \phi(\x_{s_l[i]}^\mathcal{S}\rangle)- \langle\gamma_l, \phi(\x_{t_l[i]}^\mathcal{T})\rangle|^2
\end{equation}}
where $\gamma_l\in\mathbb{S}^{f-1}$ is uniformly drawn random sample from the unit $f$-dimensional ball $\mathbb{S}^{f-1}$, and  $s_l[i]$ and $t_l[i]$ are the sorted indices of $\{\gamma_l\cdot\phi(\x_i)\}_{i=1}^M$ for source and target domains, respectively.
Note that for a fixed dimension $d$, Monte Carlo  approximation error is proportional to $O(\frac{1}{\sqrt{L}})$.
We utilize the SWD as the discrepancy measure between the   probability distributions to match them in the embedding space. Next, we discuss a major deficiency in Eq.~\eqref{eq:mainPrMatch} and our remedy to tackle it.
We utilize the SWD as the discrepancy measure between the   probability densities, $p_{\mathcal{S}}(\phi(\bm{x}_{\mathcal{S}})|C_j)\big)$ and $p_{\mathcal{T}}(\phi(\bm{x}_{\mathcal{T}})|C_j)\big)$. 

\subsection{Class-conditional Alignment of Distributions}

A main shortcoming of Eq.~\eqref{eq:mainPrMatch} is that minimizing the discrepancy between $p_\mathcal{S}(\phi(\bm{X}_\mathcal{S}))$ and $p_\mathcal{T}(\phi(\bm{X}_\mathcal{T}))$ does not guarantee semantic consistency between the two domains. To clarify this point, consider the source and target domains to be images corresponding to printed digits and handwritten digits. While the feature distributions in the embedding space could have low discrepancy, the classes might not be correctly aligned in this space, e.g. digits from a class in the target domain could be matched to a wrong class of the source domain or, even    digits from multiple classes in the target domain could be matched to the cluster of a single digit of the source domain.   In such cases, the source classifier will not generalize well on the target domain. In other words, the shared embedding space, $\mathcal{Z}$, might not be a semantically meaningful space for the target domain if we solely minimize SWD between $p_\mathcal{S}(\phi(\bm{X}_\mathcal{S}))$ and $p_\mathcal{T}(\phi(\bm{X}_\mathcal{T}))$. To solve this challenge, the encoder function should be learned such that the   class-conditioned probabilities of both domains in the embedding space are similar, i.e. $p_{\mathcal{S}}(\phi(\bm{x}_{\mathcal{S}})|C_j)\approx p_{\mathcal{T}}(\phi(\bm{x}_{\mathcal{T}})|C_j)$, where $C_j$ denotes a particular class. Given this, we can mitigate the class matching problem by  using an adapted version  of Eq.~\eqref{eq:mainPrMatch} as:
\begin{equation}
\begin{split}
\min_{\bm{v},\bm{w}}& \sum_{i=1}^N \mathcal{L}\big(h_{\bm{w}}(\phi_{\bm{v}}(\bm{x}_i^s)),\bm{y}_i^s\big)\\&+\lambda \sum_{j=1}^k D\big(p_{\mathcal{S}}(\phi_{\bm{v}}(\bm{x}_{\mathcal{S}})|C_j),p_{\mathcal{T}}(\phi_{\bm{v}}(\bm{x}_{\mathcal{T}})|C_j)\big),
\end{split}
\label{eq:mainPrMatchconditional}
\end{equation} 
where   discrepancy between distributions is minimized conditioned on classes, to enforce semantic alignment in the embedding space. Solving Eq.~\eqref{eq:mainPrMatchconditional}, however, is not tractable as the labels for the target domain are not available and the conditional distribution, $p_{\mathcal{T}}(\phi(\bm{x}_{\mathcal{T}})|C_j)\big)$, is not known.  

 \begin{algorithm}[tb!]
\caption{$\mathrm{DACAD}\left (L,\eta,\lambda  \right)$\label{CADalgorithm}} 
 {\small
\begin{algorithmic}[1]
\State \textbf{Input:} data $\mathcal{D}_{\mathcal{S}}=(\bm{X}_{\mathcal{S}},  \bm{Y}_{\mathcal{S}}); \mathcal{D}_{\mathcal{T}}=(\bm{X}_{\mathcal{S}})$,
\State \textbf{Pre-training}: 
\State \hspace{4mm} $\hat{ \theta}_0=(\bm{w}_0,\bm{v}_0) =\arg\min_{\theta}\sum_i \mathcal{L}(f_{\theta}(\bm{x}_i^s),\bm{y}_i^s)$
\For{$itr = 1,\ldots, ITR$ }
\State $\mathcal{D}_{\mathcal{PL}} = 
\{(\bm{x}_i^t,\hat{\bm{y}}_i^t)|\hat{\bm{y}}_i^t= f_{\hat{ \theta}}(\bm{x}_i^t),  p( \hat{\bm{y}}_i^t|\bm{x}_i^t)>\tau \}$
\For{ $alt = 1,\ldots, ALT$ }
\State \textbf{Update} encoder parameters using pseudo-labels:
\State \hspace{4mm} $ \hat{\bm{v}} =   \sum_{j} D\big(p_{\mathcal{S}}(\phi_{\bm{v}}(\bm{x}_{\mathcal{S}})|C_j),p_{\mathcal{SL}}(\phi_{\bm{v}}(\bm{x}_{\mathcal{T}})|C_j)\big)  $

\State \textbf{Update} entire model:
\State \hspace{4mm}  $ \hat{\bm{v}},\hat{\bm{w}} = \arg\min_{\bm{w},\bm{v}} \sum_{i=1}^N \mathcal{L}\big(h_{\bm{w}}(\phi_{\hat\hat{\bm{v}}}(\bm{x}_i^s)),\bm{y}_i^s\big)$
\EndFor
\EndFor
\end{algorithmic}}
\end{algorithm} 
 
\begin{figure*}[t!]
    \centering
    \includegraphics[width=\linewidth]{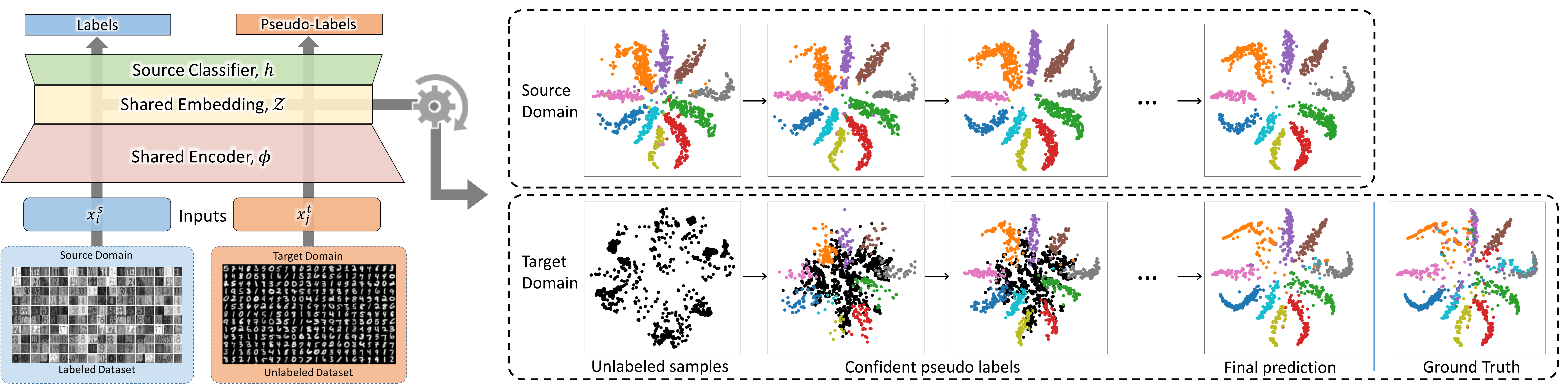}
         \caption{The high-level system architecture, shown on the left, illustrates the data paths used during UDA training. On the right,  t\_SNE visualizations  demonstrate how the embedding space evolves during training for the $\mathcal{S}\rightarrow\mathcal{U}$ task. In the target domain, colored points are examples with assigned pseudo-labels, which increase in number with the confidence of the classifier. }
         \label{fig:LL}
\end{figure*}

To tackle the above issue, we compute a surrogate of the objective in Eq.~\eqref{eq:mainPrMatchconditional}. Our idea is to approximate $p_{\mathcal{T}}(\phi(\bm{x}_{\mathcal{T}})|C_j)\big)$
by generating pseudo-labels for the target data points. The pseudo-labels are obtained from the source classifier prediction, but only for the portion of target data points that the the source classifier provides confident prediction.  
More specifically, we solve Eq.~\eqref{eq:mainPrMatchconditional} in incremental gradient descent iterations. In particular, we first initialize the classifier network by training it on the source data. 
We then
alternate between optimizing the classification loss for the source data and SWD loss term at each iteration.
At each iteration, we pass the target domain data points into  the  classifier learned on the source data and analyze the  label probability distribution on the softmax layer of the classifier. We choose a threshold $\tau$ and assign pseudo-labels only to those target data points that the classifier predicts the pseudo-labels with high confidence, i.e.   $p(\bm{y}_i|\bm{x}_i^t)>\tau$. Since the source and the target domains are related, it is sensible that the source classifier can classify a subset of target data points correctly and with high confidence.  We use these data points to approximate $p_{\mathcal{T}}(\phi(\bm{x}_{\mathcal{T}})|C_j)\big)$ in Eq.~\eqref{eq:mainPrMatchconditional} and update the encoder parameters, $\bm{v}$, accordingly.
In our empirical experiments, we have observed that because the domains are related, as more optimization iterations are performed, the number of data points with confident pseudo-labels increases and our approximation for Eq.~\eqref{eq:mainPrMatchconditional} improves and becomes more stable, enforcing the source and the target distributions to align class conditionally in the embedding space. As a side benefit, since we math the distributions class-conditionally, a problem similar to mode collapse is unlikely to occur. Figure~\ref{fig:LL}  visualizes this process  using real data. Our proposed framework, named Domain Adaptation with Conditional Alignment of Distributions (DACAD) is summarized in Algorithm~\ref{CADalgorithm}.

\section{Experimental Validation}
\label{sec:resultsSWDTL}
 
We evaluate our algorithm    using standard benchmark  UDA   tasks and compare  against  several UDA methods. 

\textit{\bf Datasets:} We investigate the empirical performance of our proposed method on five commonly used benchmark datasets in UDA, namely: MNIST ($\mathcal{M}$)~\cite{lecun1990handwritten},
USPS ($\mathcal{U}$)~\cite{lecun1995comparison}, 
Street View House Numbers, i.e., SVHN ($\mathcal{S}$), CIFAR ($\mathcal{CI}$), and STL ($\mathcal{ST}$). The first three datasets are 10 class (0-9) digit classification datasets. MNIST and USPS are collection of hand written digits whereas SVHN is a collection of real world RGB images of house numbers.  STL and CIFAR contain RGB images that share 9 object classes: \emph{airplane, car, bird, cat, deer, dog, horse, ship, and truck}.  For the digit datasets, while six domain adaptation problems can be defined among these   datasets, prior works often consider four of these six cases, as knowledge transfer from simple MNIST and USPS datasets to a more challenging SVHN domain does not seem to be tractable.  Following the literature, we use 2000 randomly selected images from MNIST and 1800 images from USPS in our experiments for the case of $\mathcal{U}\rightarrow \mathcal{M}$ and $\mathcal{S}\rightarrow\mathcal{M}$~\cite{motiian2017few}. In the remaining cases, we used full datasets. All datasets have their images scaled to $32\times 32$ pixels and the SVHN images are converted to grayscale as the encoder network is shared between the domains. CIFAR and STL maintain their RGB components. We report the target classification accuracy across the tasks.

\textit{\bf Pre-training}: Our experiments involve a pre-training stage to initialize the encoder and the classifier networks solely using the source data. This is an essential step because the combined deep network can generate confident pseudo-labels on the target domain only if initially trained on the related source domain. In other words, this initially learned network   can be served as a naive model on the target domain.  We then boost the performance on the target domain using our proposed algorithm, demonstrating that our algorithm is indeed effective for transferring knowledge. 
Doing so, we investigate a less-explored issue in the UDA literature. Different UDA approaches use considerably different networks, both in terms of complexity, e.g. number of layers and convolution filters, and the structure, e.g. using an auto-encoder. Consequently, it is ambiguous  whether  performance of a particular UDA algorithm is due to successful knowledge transfer from the source domain or just a good baseline network that performs well on the target domain even without considerable knowledge transfer from the source domain. To highlight that our algorithm can indeed transfer knowledge, we use  two different network architectures: DRCN~\cite{ghifary2016deep} and VGG~\cite{simonyan2014very}. 
We then show that our algorithm can effectively boost base-line performance (statistically significant) regardless of the underlying network. In most of the  domain adaptation tasks, we demonstrate that this boost indeed stems from transferring knowledge from the source domain. In our experiments we used Adam optimizer ~\cite{kingma2014adam}  and set the pseudo-labeling  threshold to $tr = 0.99$.

 \begin{table*}[t!]
 \centering 
{\small
\begin{tabular}{lc|cccccc}   
\multicolumn{2}{c}{Method}    & $\mathcal{M}\rightarrow\mathcal{U}$ & $\mathcal{U}\rightarrow\mathcal{M}$ & $\mathcal{S}\rightarrow\mathcal{M}$ &$\mathcal{S}\rightarrow\mathcal{U}$ &$\mathcal{ST}\rightarrow\mathcal{CI}$ &$\mathcal{CI}\rightarrow \mathcal{ST}$\\
\hline
\multicolumn{2}{c|}{GtA}& 92.8$\pm$0.9	&	90.8$\pm$1.3	&	92.4$\pm$0.9 &- &-&-\\
\multicolumn{2}{c|}{CoGAN}& 91.2$\pm$0.8&89.1$\pm$0.8&-&-&-&-   \\ 
\multicolumn{2}{c|}{ADDA}& 89.4$\pm$0.2&90.1$\pm$0.8&76.0$\pm$1.8&-&-&-\\
\multicolumn{2}{c|}{CyCADA}&  95.6$\pm$0.2$^{\dagger}$ & 96.5$\pm$0.1$^{\dagger}$ & 90.4$\pm$0.4&-&-&- \\  
\multicolumn{2}{c|}{I2I-Adapt}& 92.1$^{\dagger}$ & 87.2$^{\dagger}$ & 80.3&-&-&-   \\
\multicolumn{2}{c|}{FADA }& 89.1 & 81.1 & 72.8& 78.3&-&-	 \\  
\hline
\multicolumn{2}{c|}{RevGrad}&	 77.1$\pm$1.8$^{\ddagger}$	&	73.0$\pm$2.0$^{\ddagger}$	&	73.9  &-&-&-   \\ 
\multicolumn{2}{c|}{DRCN}&	 91.8$\pm$0.1$^{\dagger}$	&	73.7$\pm$0.04$^{\dagger}$	&	82.0$\pm$0.2 &- & 58.9$\pm$0.1 & 66.4$\pm$0.1  \\
\multicolumn{2}{c|}{AUDA}& - & - & 86.0 & -&-&- \\ 
\multicolumn{2}{c|}{OPDA}& 70.0 & 60.2 &-&-&-&- \\ 
\multicolumn{2}{c|}{MML}& 77.9$^{\dagger}$ & 60.5$^{\dagger}$ &62.9 &-&-&- \\ 
\hline
\hline
\multicolumn{2}{c|}{Target (FS)}&	96.8$\pm$0.2 &	98.5$\pm$0.2 &	98.5$\pm$0.2 & 96.8$\pm$0.2 & 81.5$\pm$1.0 & 64.8$\pm$1.7 \\
\hline
\multicolumn{2}{c|}{VGG}&	 90.1$\pm$2.6	&	80.2$\pm$5.7	&	67.3$\pm$2.6 & 66.7$\pm$2.7 & \cellcolor{LightRed} 53.8$\pm$1.4 & 63.4$\pm$1.2 \\
\multicolumn{2}{c|}{DACAD}&	 92.4$\pm$1.2	&	91.1$\pm$3.0	&	80.0$\pm$2.7 & 79.6$\pm$3.3  & \cellcolor{LightRed} 54.4$\pm$1.9 & 66.5$\pm$1.0 \\
\hline
\multicolumn{2}{c|}{DACAD}&	 93.6$\pm$1.0	&	95.8$\pm$1.3	&	82.0$\pm$3.4 & 78.0$\pm$2.0  & \cellcolor{LightRed} 44.4$\pm$0.4 & 65.7$\pm$1.0 \\
\hline
\multicolumn{2}{c|}{DRCN (Ours)}&	 88.6$\pm$1.3	&	89.6$\pm$1.3	&	74.3$\pm$2.8 & 54.9$\pm$1.8  & 50.0$\pm$1.5 & 64.2$\pm$1.7 \\
\multicolumn{2}{c|}{DACAD}&	 94.5$\pm$0.7	&	97.7$\pm$0.3	&	88.2$\pm$2.8 & 82.6$\pm$2.9  & 55.0$\pm$1.4 & 65.9$\pm$1.4 \\
\end{tabular}}
\caption{ Classification accuracy for UDA between MINIST, USPS, SVHN, CIFAR, and STL datasets. $^{\dagger}$ indicates use of full MNIST and USPS datasets as opposed to the subset described in the paper. $^{\ddagger}$ indicates results from reimplementation in ~\cite{tzeng2017adversarial}. }
\label{table:tabDA1}
\end{table*}



 \textit{\bf Data Augmentation:} Following the literature, we use data augmentation to create additional training data by applying reasonable transformations to input data in an effort to improve generalization~\cite{simard2003augmentation}. 
Confirming the reported result in \cite{ghifary2016deep}, we also found that geometric transformations and noise, applied to appropriate inputs, greatly improves performance and transferability
 of the source model to the target data. Data augmentation can help to reduce the domain shift between the two domains. The augmentations in this work are limited to translation, rotation, skew, zoom, Gaussian noise, Binomial noise, and inverted pixels.
 

\subsection{Results}
  
Figure~\ref{fig:LL} demonstrates how our algorithm successfully learns an embedding with class-conditional alignment of distributions of both domains. 
This figure presents the two-dimensional t\_SNE visualization of the source  and target domain data points in the shared embedding space for the $\mathcal{S}\rightarrow \mathcal{U}$ task. The horizontal axis demonstrates the optimization iterations where each cell presents data visualization after a particular optimization iteration is performed. The top sub-figures visualize the source data points, where each color represents a particular class. The bottom sub-figures visualize the target data points, where the colored data points represent the pseudo-labeled data points at each iteration and the black points represent the rest of the target domain data points. We can see that, due to pre-training initialization, the embedding space is discriminative for the source domain in the beginning, but the target distribution differs from the source distributions. However, the classifier is confident about a portion of target data points. As more optimization iterations are performed, since the network becomes a better classifier for the target domain, the number of the target pseudo-labeled data points increase, improving our approximate of Eq.~\ref{eq:mainPrMatchconditional}. As a result, the discrepancy between the two distributions progressively decreases. Over time, our algorithm learns a shared embedding which is discriminative for both domains, making pseudo-labels a good prediction for the original labels, bottom, right-most sub-figure. This result empirically validates 
applicability of our algorithm to address UDA.

 We also compare our results against several recent UDA algorithms in Table~\ref{table:tabDA1}.
 In particular, we compare against the recent adversarial learning algorithms: Generate to Adapt (GtA)~\cite{sankaranarayanan2017generate}, CoGAN~\cite{liu2016coupled}, ADDA~\cite{tzeng2017adversarial},   CyCADA~\cite{hoffman2017cycada}, and I2I-Adapt~\cite{murez2018image}. We also include FADA~\cite{motiian2017few}, which is originally a few-shot learning technique. 
 For FADA, we list the reported one-shot accuracy, which is very close to the UDA setting (but it is arguably a simpler problem). Additionally, we have included results for RevGrad~\cite{ganin2014unsupervised}, DRCN~\cite{ghifary2016deep}, AUDA~\cite{saito2017asymmetric}, OPDA~\cite{courty2017optimal}, MML~\cite{seguy2017large}. The latter methods  are   similar to our method because these methods learn an embedding space to couple the domains. OPDA and MML  are more similar as they  match distributions explicitly in the learned embedding. Finally, we have included the performance of fully-supervised (FS) learning on the target domain as an upper-bound for UDA. In our own results, we  include the baseline target performance that we obtain by naively employing a DRCN network as well as target performance from VGG  network that are learned solely on the source domain. We notice that in Table~\ref{table:tabDA1},  our baseline performance is better than some of the UDA algorithms for some tasks. This is a very crucial observation as it demonstrates that, in some cases, a trained deep network with good data augmentation can extract domain invariant features that make domain adaptation feasible even without any further transfer learning procedure. The last row demonstrates  that our method is effective in transferring knowledge to boost the baseline performance. In other words, Table~\ref{table:tabDA1} serves as an ablation study to demonstrate that   that effectiveness of our algorithm stems from successful cross-domain knowledge transfer.  We can   see that our algorithm   leads to near- or the state-of-the-art performance across the tasks. Additionally, an important observation is that our method significantly outperforms the methods that match distributions directly and is competent against methods that use adversarial learning. This can be explained as the result of matching distributions class-conditionally and suggests our second contribution can potentially boost performance of these methods. Finally, we note that our proposed method provide a statistically significant boost in all but two of the cases (shown in gray in Table~\ref{table:tabDA1}).

\section{Conclusions and Discussion} 
We developed a new UDA algorithm based on learning a domain-invariant embedding space. We  map   data points from   two related domains to the embedding space such that discrepancy between the transformed distributions is minimized.   We used the sliced Wasserstein distance metric as a measure to match the distributions in the embedding space. As a result, our method is computationally more efficient.  Additionally, we matched distributions class-conditionally by assigning pseudo-labels to the target domain data. As a result, our method is more robust and outperforms prior UDA methods that match distributions directly. We provided
 experimental validations to demonstrate that our method is competent against SOA recent UDA methods.

\end{document}